\documentclass[11pt]{article}

\usepackage[margin=1in]{geometry}
\usepackage{amsmath,amssymb,amsfonts}
\usepackage{graphicx}
\usepackage{xcolor}
\usepackage{booktabs}
\usepackage{multirow}
\usepackage{url}
\usepackage{cite}
\usepackage{float}
\usepackage[linesnumbered,ruled,vlined]{algorithm2e}

\title{Hierarchical Flow Decomposition for Turning Movement Prediction at Signalized Intersections}

\author{
Md Atiqur Rahman Mallick \and
Kamrul Hasan \and
Pulock Das \and
Liang Hong \and
S M Shazzad Rassel \\
\\
Department of Electrical and Computer Engineering\\
Tennessee State University, Nashville, TN, USA\\
\texttt{\{mmallick,mhasan1,pdas,lhong,srassel\}@tnstate.edu}
}

\date{}

\begin{document}

\maketitle

\begin{center}
{\small
This is the author’s accepted manuscript of a paper accepted for publication in the Proceedings of IEEE SoutheastCon 2026. The final version will be available via IEEE Xplore.
}
\end{center}

\begin{abstract}

Accurate prediction of intersection turning movements is essential for adaptive signal control but remains difficult due to the high volatility of directional flows. This study proposes HFD-TM (Hierarchical Flow-Decomposition for Turning Movement Prediction), a hierarchical deep learning framework that predicts turning movements by first forecasting corridor through-movements and then expanding these predictions to individual turning streams. This design is motivated by empirical traffic structure, where corridor flows account for 65.1\% of total volume, exhibit lower volatility than turning movements, and explain 35.5\% of turning-movement variance. A physics-informed loss function enforces flow conservation to maintain structural consistency. Evaluated on six months of 15-minute intervals of LiDAR (Light Detection and Ranging) data from a six-intersection corridor in Nashville, Tennessee, HFD-TM achieves a mean absolute error of 2.49 vehicles per interval, reducing MAE by 5.7\% compared to a Transformer and by 27.0\% compared to a GRU (Gated Recurrent Unit). Ablation results show that hierarchical decomposition provides the largest performance gain, while training time is 12.8× lower than DCRNN(Diffusion Convolutional Recurrent Neural Network), demonstrating suitability for real-time traffic applications.

\end{abstract}


\section{Introduction}

Accurate forecasting of intersection turning movements is pivotal for optimizing urban traffic signal control and enabling proactive Intelligent Transportation Systems (ITS) \cite{Zhang2024RealTime,Watson2024ShortTermTMF,Li2024IntersectionSurvey}. Despite advances in deep learning, modeling these fine-grained dynamics remains challenging due to the inherent stochasticity and volatility of traffic flow at the intersection level. Unlike smoother corridor-level metrics, directional turning movements fluctuate rapidly due to signal phases and upstream disturbances \cite{Afandizadeh2024DLReview,tedjopurnomo2023survey,li2025traffic}. This creates a bottleneck for real-world deployment, as traditional methods often fail to capture complex spatiotemporal dependencies. Classical statistical models like ARIMA assume linearity, rendering them ineffective for non-linear traffic dynamics, while standard machine learning approaches like Support Vector Regression are constrained by shallow architectures that cannot model long-range temporal dependencies \cite{Ali2025SurveyDL,Han2025}.

Recent deep learning adoptions, such as Long Short-Term Memory (LSTM) and Gated Recurrent Units (GRU), have improved sequence forecasting but often treat traffic streams as isolated time series, neglecting the structural relationship between corridor throughput and local turning movements \cite{toba2025long,Ren2025VMDGAT}. While advanced Graph Neural Networks (GNNs) and Transformer-based models attempt to capture network-level dependencies, they frequently suffer from excessive computational complexity and a ``black-box'' nature that ignores physical constraints like flow conservation \cite{Ahmed2024EnhancementTraffic,Pan2025PhysicsGuided,xu2020spatial}. These complex hybrid architectures are difficult to train and often fail to explicitly model the hierarchical interaction between upstream volumes and downstream turning fractions \cite{wang2025hybrid,zhang2025spatial,kong2024spatio}, necessitating a more robust, physics-informed approach that balances computational efficiency with structural consistency \cite{usama2022physics}.

In this study, we propose HFD-TM, a deep learning framework for intersection turning-movement prediction that hierarchically separates corridor-level through flows from intersection-level turning streams while enforcing flow conservation through a physics-informed loss. Rather than predicting each turning movement directly, HFD-TM reformulates the task as a hierarchical conditional problem, in which volatile turning movements are modeled as realizations conditioned on lower-variance corridor through-flows that maintain structural consistency, and computationally efficient forecasting suitable for real-time traffic signal controls\cite{wei2025hierarchical}.

The key contributions of this work are summarized as follows:
\begin{itemize}
    \item \textbf{Hierarchical modeling:} A hierarchical framework that forecasts corridor-level through movements before estimating intersection turning movements.
    \item \textbf{Learned turning expansion:} A turn-movement expansion module that combines corridor predictions with time-of-day embeddings to generate full turning streams.
    \item \textbf{Constrained refinement:} A refinement stage with residual correction and zero-movement masking to enforce temporal continuity and geometric feasibility.
\end{itemize}

The remainder of this paper is organized as follows: Section II reviews related work; Section III presents the proposed HFD-TM methodology; Section IV describes the experimental setup and results; Section V discusses comparative performance; and Section VI concludes the paper.

\section{Related Work}

Early methodologies for estimating turning movements relied heavily on static analytical techniques, such as iterative proportional fitting, to resolve flow distributions from limited boundary constraints \cite{Zhang2024RealTime}. While effective for long-term planning, these approaches struggle to accommodate the stochastic volatility inherent in real-time operations. To address temporal dynamics, researchers applied classical time-series models, including Autoregressive Integrated Moving Average (ARIMA) and Kalman filtering \cite{Afandizadeh2024DLReview}. However, the fundamental assumption of linearity within these frameworks limits their applicability to the non-linear flow patterns observed at signalized intersections \cite{Ali2025SurveyDL}. The constraints of parametric models necessitated a shift toward non-parametric machine learning. Artificial Neural Networks (ANNs) and Support Vector Regression (SVR) emerged as effective alternatives, offering the capability to approximate complex functions without rigid theoretical assumptions \cite{Ali2025SurveyDL}. Despite improving predictive accuracy, these architectures often process traffic data as isolated vectors, failing to capture long-range temporal dependencies \cite{li2025traffic}.

Enhancing the robustness of deep learning methods in the face of stochastic traffic dynamics remains a crucial area of investigation \cite{Li2024IntersectionSurvey}. Recurrent architectures, specifically bi-directional LSTMs and Spatio-Temporal Transformers, have shown utility in sequence modeling by analyzing temporal contexts in both forward and reverse directions \cite{Han2025,xu2020spatial}. However, purely data-driven sequence models often neglect fundamental spatial constraints. To address these challenges, hierarchical spatial–temporal graph convolutional networks (ST-GCNs) and federated learning frameworks have been proposed to model complex topological dependencies in urban traffic networks \cite{Pan2025PhysicsGuided,Feng2024FedTFP}. While effective in capturing citywide propagation, ST-GCN-based approaches face substantial computational overhead and difficulty generalizing across heterogeneous geometries \cite{Ahmed2024EnhancementTraffic}. Similarly, adaptive spatiotemporal feature fusion networks have been introduced to improve intersection-level prediction by dynamically weighting input features \cite{gong2024parallel,kong2024spatio}. However, such approaches often struggle to ensure structural consistency, as predicted turning fractions may fail to sum to the entering volume \cite{Liu2025PNNs4}. Attention-based models using historical traffic sequences have further improved short-term forecasting performance \cite{Ren2025VMDGAT}. Despite these advances, existing methods generally do not explicitly model the hierarchical interaction between upstream corridor flows and downstream turning movements \cite{wang2025hybrid,zhang2025spatial,wei2025hierarchical}, nor do they fully exploit physics-informed constraints for regularization \cite{usama2022physics,li2025traffic}. To address these computational and structural limitations, the proposed HFD-TM framework adopts a hierarchical decomposition strategy integrated with a physics-informed conservation objective.

\section{Methodology}

\subsection{Proposed HFD-TM Framework}

The proposed Hierarchical Flow-Decomposition for Turning Movements (HFD-TM) framework 
\begin{figure}[t]
    \centering
    \includegraphics[width=\linewidth]{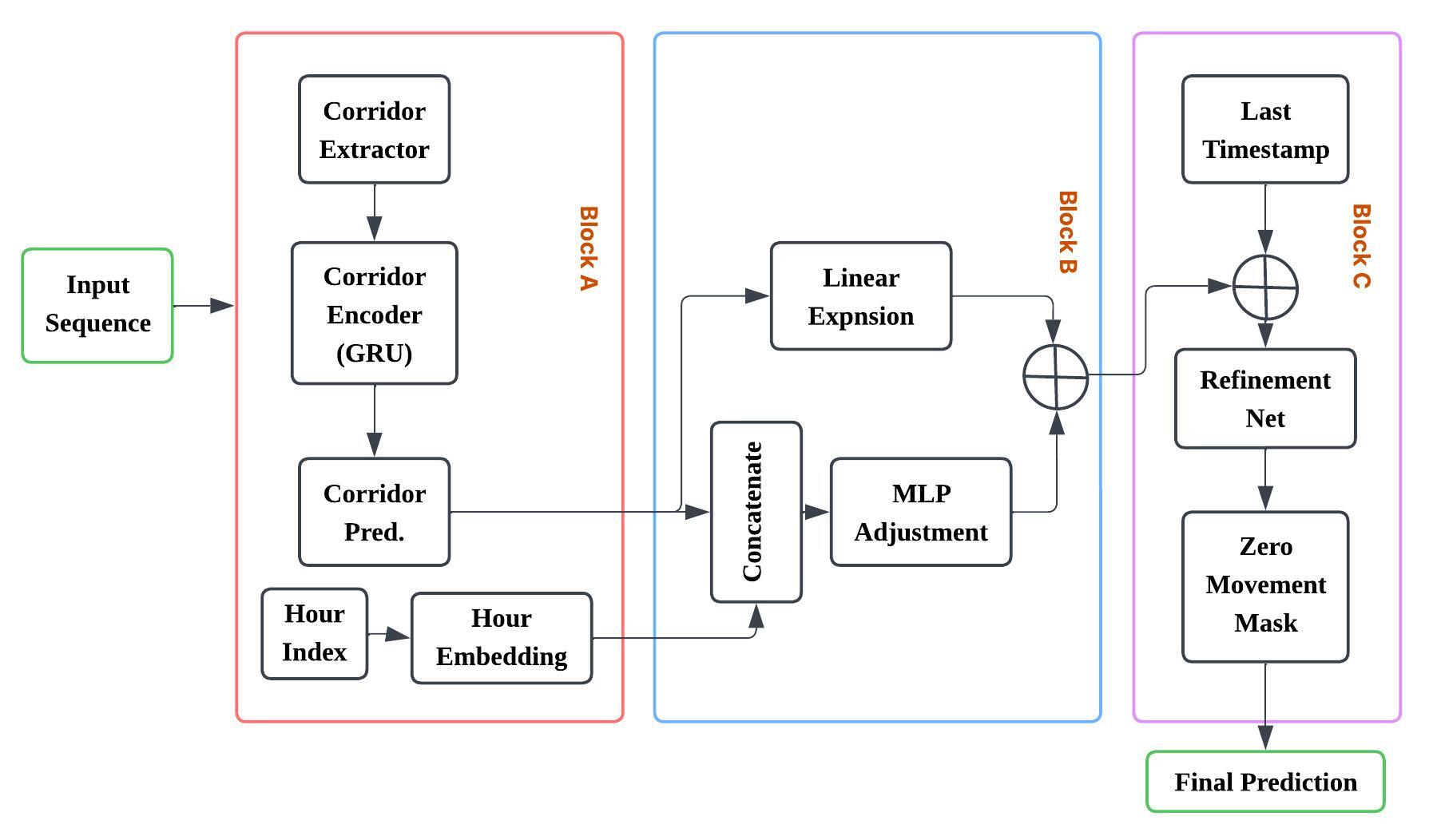}
    \caption{System-level overview of the proposed HFD-TM framework. Block A encodes corridor-level through flows from historical inputs, Block B expands corridor predictions into turning movements using contextual information, and Block C refines predictions with residual correction and zero-movement masking.}
    \label{fig:placeholder}
\end{figure}
predicts intersection turning-movement counts by decomposing the task into two stages: 
(i) forecasting dominant corridor-level through movements and 
(ii) expanding these forecasts to obtain full turning-movement streams. 
This design reflects the empirical structure of corridor traffic, where through 
movements account for the majority of volume and exhibit lower temporal variability 
than individual turning movements.

Let the normalized input sequence be
\begin{equation}
\mathbf{X} \in \mathbb{R}^{T \times N},
\end{equation}
where $T$ denotes the input sequence length and $N$ is the total number of movement 
streams across the corridor.

Let $\mathcal{C} \subset \{1,\ldots,N\}$ denote the index set corresponding to corridor 
(through) movements. The corridor-level input is defined as
\begin{equation}
\mathbf{X}_c = \mathbf{X}[:, \mathcal{C}] \in \mathbb{R}^{T \times N_c},
\end{equation}
where $N_c = |\mathcal{C}|$.

The HFD-TM framework generates predictions in a hierarchical manner. 
It first produces a corridor-level prediction
\begin{equation}
\hat{\mathbf{y}}_c \in \mathbb{R}^{N_c},
\end{equation}
which represents the predicted through-movement volumes along the corridor. 
These corridor predictions are subsequently expanded through a learned 
turn-movement prediction module to obtain the full turning-movement prediction
\begin{equation}
\hat{\mathbf{y}} \in \mathbb{R}^{N}.
\end{equation}

\subsection{Turning-Movement Expansion via Ratio Prediction}
The expansion from corridor predictions to full movement predictions is implemented using a TurnRatioPredictor module. This module combines corridor predictions with a learned hour-of-day embedding to generate an additive correction on top of a linear expansion.

Let the hour-of-day input be
\begin{equation}
h \in \{0, \ldots, 23\},
\end{equation}
provided as a LongTensor. The hour embedding is defined as
\begin{equation}
\mathbf{e}_h = \mathrm{Embed}(h) \in \mathbb{R}^{H},
\end{equation}
where $H$ is the hidden dimension (set to 64 in this work).

A linear expansion of corridor predictions is first computed as
\begin{equation}
\hat{\mathbf{y}}_{\mathrm{expand}} = \mathbf{W}_e \hat{\mathbf{y}}_c + \mathbf{b}_e \in \mathbb{R}^{N}.
\end{equation}

The input to the multilayer perceptron (MLP) is the concatenation of corridor predictions and the hour embedding:
\begin{equation}
\mathbf{z} = [\hat{\mathbf{y}}_c, \mathbf{e}_h] \in \mathbb{R}^{N_c + H}.
\end{equation}

The MLP produces an additive adjustment:
\begin{equation}
\Delta \hat{\mathbf{y}} = f_{\mathrm{MLP}}(\mathbf{z}) \in \mathbb{R}^{N}.
\end{equation}

The expanded full-movement prediction is then given by
\begin{equation}
\hat{\mathbf{y}}_{\mathrm{all}} = \hat{\mathbf{y}}_{\mathrm{expand}} + \Delta \hat{\mathbf{y}}.
\end{equation}

\subsection{Residual Refinement and Zero-Movement Masking}
After expansion, a residual connection from the most recent observed full-movement vector is applied. Let
\begin{equation}
\mathbf{x}_T \in \mathbb{R}^{N}
\end{equation}
denote the last time step of the input sequence. The residual weight was chosen empirically based on validation performance, with stable results observed for values between 0.0 and 0.5. A fixed value of 0.3 was selected to balance recent observations and model predictions while avoiding additional parameters and training complexity. This choice ensures that the residual term provides temporal smoothing without allowing persistence to dominate the hierarchical prediction.

\begin{equation}
\hat{\mathbf{y}}_{\mathrm{res}} = \hat{\mathbf{y}}_{\mathrm{all}} + 0.3 \, \mathbf{x}_T.
\end{equation}

A lightweight refinement network then produces a residual correction:
\begin{equation}
\hat{\mathbf{y}}_{\mathrm{ref}} = f_{\mathrm{refine}}(\hat{\mathbf{y}}_{\mathrm{res}}) + \hat{\mathbf{y}}_{\mathrm{res}}.
\end{equation}

The refinement network is intentionally lightweight (two fully connected layers) to preserve computational efficiency and avoid overfitting. Finally, a zero-movement mask, are identified based on intersection geometry and lane configuration and remain fixed during training,
\begin{equation}
\mathbf{m} \in \{0,1\}^{N}
\end{equation}
is applied to enforce physically implausible zero movements:
\begin{equation}
\hat{\mathbf{y}} = \hat{\mathbf{y}}_{\mathrm{ref}} \odot \mathbf{m}.
\end{equation}

The mask $\mathbf{m}$ is implemented as a registered buffer, assigning zeros to indices corresponding to structurally infeasible movements, where $\odot$ denotes element-wise multiplication.

\begin{algorithm}[t]
\LinesNotNumbered
\IncMargin{1.0em}
\caption{Proposed HFD-TM Based Turning Movement Prediction Procedure}
\label{alg:hfd_tm_short}
\KwData{Normalized sequences $\mathbf{X}$, targets $\mathbf{y}$, hour indices $h$; corridor indices $\mathcal{C}$; active indices $\mathcal{A}$; intersection groups $\{\mathcal{G}_k\}_{k=1}^{K}$}
\KwResult{Predicted turning movements $\hat{\mathbf{y}}$; trained parameters $\Theta^*$}
\SetKwFunction{Main}{Main}
\SetKwProg{Fn}{Function}{:}{}
\Fn{\Main{$(\mathbf{X}, \mathbf{y}, h)$}}{
\textbf{Step 1: Initialize Model and Optimizer}\;
\Indp
Initialize HFD-TM $f_{\Theta}$ (GRU encoder, ratio predictor, refine block, zero-mask)\;
Initialize Adam optimizer ($lr{=}0.001$, $wd{=}5{\times}10^{-4}$) and LR scheduler\;
\Indm
\BlankLine
\textbf{Step 2: Train the HFD-TM Model}\;
\Indp
\For{epoch $=1$ to $E_{\max}$}{
    \ForEach{mini-batch $(\mathbf{X}_b,\mathbf{y}_b,h_b)$}{
        \textbf{Step 2.1:} Corridor Encoding\;
        \Indp
        $\mathbf{X}_{c} \leftarrow \mathbf{X}_b[:,:, \mathcal{C}]$\;
        $\hat{\mathbf{y}}_{c} \leftarrow \text{CorridorEncoder}(\mathbf{X}_{c})$\;
        \Indm
        \BlankLine
        \textbf{Step 2.2:} Turning Expansion and Refinement\;
        \Indp
        $\hat{\mathbf{y}} \leftarrow \text{TurnRatioPredictor}(\hat{\mathbf{y}}_{c}, h_b)$\;
        $\hat{\mathbf{y}} \leftarrow \text{Refine}(\hat{\mathbf{y}}, \mathbf{X}_b)$\;
        $\hat{\mathbf{y}} \leftarrow \hat{\mathbf{y}} \odot \text{zero\_mask}$\;
        \Indm
        \BlankLine
        \textbf{Step 2.3:} Hierarchical Loss and Update\;
        \Indp
        $\mathcal{L} \leftarrow \text{MSE}(\hat{\mathbf{y}}[:,\mathcal{A}], \mathbf{y}_b[:,\mathcal{A}])
        + \lambda_{\text{corr}}\text{MSE}(\hat{\mathbf{y}}[:,\mathcal{C}], \mathbf{y}_b[:,\mathcal{C}])
        + \lambda_{\text{cons}}\mathcal{L}_{\text{cons}}(\hat{\mathbf{y}},\mathbf{y}_b,\{\mathcal{G}_k\})$\;
        Backpropagate $\mathcal{L}$ and update $\Theta$\;
        \Indm
    }
    Update LR scheduler and apply early stopping on validation MAE\;
}
\Indm
Save best checkpoint $\Theta^*$\;
\BlankLine
\textbf{Step 3: Inference}\;
\Indp
$\hat{\mathbf{y}} \leftarrow f_{\Theta^*}(\mathbf{X}, h)$\;
\Indm
\Return $\hat{\mathbf{y}}$\;
}
\DecMargin{1.0em}
\end{algorithm}

\subsection{Hierarchical Loss Formulation}
Training is performed using a hierarchical loss function that combines prediction accuracy, corridor emphasis, and flow conservation constraints.

Let $\mathcal{A}$ denote the index set of active movements and $\mathcal{C}$ the corridor movement indices. The active-movement mean squared error (MSE) is defined as
\begin{equation}
\mathcal{L}_{\mathrm{mse}} = \mathrm{MSE}(\hat{\mathbf{y}}[:, \mathcal{A}], \mathbf{y}[:, \mathcal{A}]).
\end{equation}
The active index set excludes structurally inactive movements to avoid biasing the loss with trivial zero flows.
A corridor-weighted MSE term is added:
\begin{equation}
\mathcal{L}_{\mathrm{corr}} = \mathrm{MSE}(\hat{\mathbf{y}}[:, \mathcal{C}], \mathbf{y}[:, \mathcal{C}]).
\end{equation}

Corridor movements are weighted separately due to their dominant contribution to total volume and their influence on downstream turning predictions. To enforce flow conservation at intersections, movements are grouped by intersection. Let $\mathcal{G}_k$ denote the index set of movements belonging to intersection $k$, and let $K$ be the total number of intersections. The conservation loss is defined as
\begin{equation}
\mathcal{L}_{\mathrm{cons}} = \frac{1}{K} \sum_{k=1}^{K}
\mathrm{MSE}
\left(
\sum_{i \in \mathcal{G}_k} \hat{y}_i,
\sum_{i \in \mathcal{G}_k} y_i
\right).
\end{equation}

The total training loss is given by
\begin{equation}
\mathcal{L} =
\mathcal{L}_{\mathrm{mse}}
+ \lambda_{\mathrm{corr}} \mathcal{L}_{\mathrm{corr}}
+ \lambda_{\mathrm{cons}} \mathcal{L}_{\mathrm{cons}}.
\end{equation}

The weighting parameters $\lambda_{\text{corr}}$ and $\lambda_{\text{cons}}$ are chosen to balance prediction accuracy, corridor fidelity, and flow conservation while maintaining stable optimization behavior.

\section{Proposed Framework Training and Optimization}

The HFD-TM framework is trained in an end-to-end supervised manner using normalized temporal input sequences, corresponding turning-movement targets, and discrete hour-of-day indices. Training strictly follows the hierarchical modeling pipeline described in Section~III and is governed by the optimization procedure summarized in Algorithm~1.

\subsection{Training Setup}

Let $(X, y, h)$ denote the normalized input sequences, ground-truth turning-movement vectors, and hour-of-day indices, respectively. Corridor movement indices $C$, active movement indices $A$, and intersection groupings $\{G_k\}_{k=1}^{K}$ are predefined and fixed throughout training. All model parameters $\Theta$, including the corridor encoder, turn-ratio predictor, residual refinement block, and the registered zero-movement mask, are jointly optimized.

The model is trained using the Adam optimizer with a learning rate of $1\times10^{-3}$ and a weight decay of $5\times10^{-4}$. A learning-rate scheduler is applied during training, and early stopping based on validation mean absolute error (MAE) is employed to prevent overfitting. The model checkpoint achieving the best validation performance is retained for inference.

\subsection{Mini-Batch Training Procedure}

During each training epoch, the dataset is processed in mini-batches. Given a mini-batch $(X_b, y_b, h_b)$, corridor-level inputs are first extracted as
\begin{equation}
X_c = X_b[:, :, C].
\end{equation}
The corridor encoder processes $X_c$ to generate corridor-level predictions $\hat{y}_c$. These predictions are subsequently passed to the turn-ratio predictor, together with the corresponding hour-of-day embeddings, to generate expanded turning-movement estimates.

The expanded predictions are further refined through a residual refinement block that incorporates a fixed residual contribution from the most recent observed full-movement vector. Finally, a registered zero-movement mask is applied to enforce physically implausible zero-flow constraints prior to loss evaluation.


\section{Experimental Evaluation}
\label{sec:experiments}

\subsection{Experimental Setup and Evaluation Metrics}
\label{sec:setup_metrics}
The proposed HFD-TM framework is evaluated using real world LiDAR based traffic data collected from a six-intersection corridor along Clarksville Pike in Nashville, Tennessee. The dataset was obtained via the Bluecity platform provided by the Nashville Department of Transportation (NDOT), which aggregates and distributes traffic measurements from deployed roadside LiDAR sensors. It contains 20,352 samples recorded at 15 minutes intervals over six months and includes 72 movement streams (12 corridor through-movements and 60 turning movements). Hour of day indices extracted from timestamps are provided as auxiliary inputs to capture diurnal traffic patterns.
\begin{table*}[t]
\centering
\caption{Baseline comparison results. Errors are reported in vehicles per 15-minute interval.}
\label{tab:baseline}

\setlength{\tabcolsep}{24pt}   
\renewcommand{\arraystretch}{1.15}

\begin{tabular}{lrrrrrr}
\toprule
Model & MAE & RMSE & Time (s) & MAE $\Delta$ & RMSE $\Delta$ \\
\midrule
GRU & 3.4071 & 8.1654 & 67.4   & +27.0\% & +36.5\% \\
LSTM & 3.5363 & 8.4549 & 44.1   & +29.7\% & +38.6\% \\
Transformer & 2.6383 & 5.7130 & 230.1 & +5.7\%  & +9.2\% \\
DCRNN & 2.6779 & 5.8697 & 3539.9 & +7.1\%  & +11.6\% \\
\textbf{HFD-TM} & \textbf{2.4873} & \textbf{5.1886} & 275.8 & -- & -- \\
\bottomrule
\end{tabular}
\end{table*}

The data are split chronologically into training, validation, and test sets using a 70\%--15\%--15\% partition to preserve temporal ordering and avoid leakage. All movement counts are normalized using min--max scaling parameters computed on the training set and applied consistently to validation and test sets.

Performance is evaluated using Mean Absolute Error (MAE) and Root Mean Squared Error (RMSE), reported in vehicles per 15-minute interval:
\begin{equation}
\mathrm{MAE} = \frac{1}{n}\sum_{i=1}^{n}\left|y_i-\hat{y}_i\right|,
\label{eq:mae}
\end{equation}
\begin{equation}
\mathrm{RMSE} = \sqrt{\frac{1}{n}\sum_{i=1}^{n}\left(y_i-\hat{y}_i\right)^2}.
\label{eq:rmse}
\end{equation}
Training time (seconds) is also recorded to assess computational efficiency.

\subsection{Baseline Comparison Results}
\label{sec:baseline_results}
HFD-TM is compared with four baseline models: GRU, LSTM, Transformer, and DCRNN. All models are trained using identical data splits and evaluated using the same metrics. Table 1 summarizes the results. HFD-TM achieves the lowest MAE (2.4873) and RMSE (5.1886), outperforming all baselines. Relative to the strongest baseline (Transformer, MAE = 2.6383), HFD-TM reduces MAE by 5.7\% and RMSE by 9.2\%. Larger gains are observed relative to GRU and LSTM, with MAE reductions of 27.0\% and 29.7\%, respectively. In addition, HFD-TM trains in 275.8 seconds, which is approximately 12.8$\times$ faster than DCRNN while achieving higher accuracy.


\subsection{Training Stability and Convergence Analysis}
\label{sec:convergence}
All models are trained using the Adam optimizer with early stopping based on validation MAE. HFD-TM converges consistently within 60 epochs. Direct comparison of raw training loss magnitudes across models is not meaningful, as the HFD-TM objective includes additional terms (corridor weighting and conservation). Accordingly, model comparison is based on test-set accuracy at the best validation checkpoint.

\subsection{Ablation Study and Component Contribution}
\label{sec:ablation}
An ablation study is conducted to assess the contribution of individual components in HFD-TM. Results are reported in Table 2
\begin{table}[t]
\centering
\caption{Ablation Study Results. $\Delta$MAE is reported relative to the full HFD-TM model.}
\label{tab:ablation}
\renewcommand{\arraystretch}{1.15}
\begin{tabular}{lrrr}
\toprule
Configuration & MAE & RMSE & $\Delta$MAE \\
\midrule
\textbf{Full Model} & \textbf{2.4873} & \textbf{5.1886} & -- \\
$-$ Hierarchy & 2.6035 & 5.7393 & +4.67\% \\
$-$ Corridor Weight & 2.5474 & 5.4205 & +2.41\% \\
$-$ Conservation & 2.5437 & 5.3142 & +2.27\% \\
\bottomrule
\end{tabular}
\end{table}
. Removing hierarchical decomposition increases MAE from 2.4873 to 2.6035 (+4.67\%), representing the largest degradation. Removing the corridor-weighting term increases MAE to 2.5474 (+2.41\%), while removing the conservation term increases MAE to 2.5437 (+2.27\%). These results confirm that all components contribute positively, with hierarchical decomposition providing the dominant benefit.

\begin{figure}
    \centering
    \includegraphics[width=0.95\linewidth]{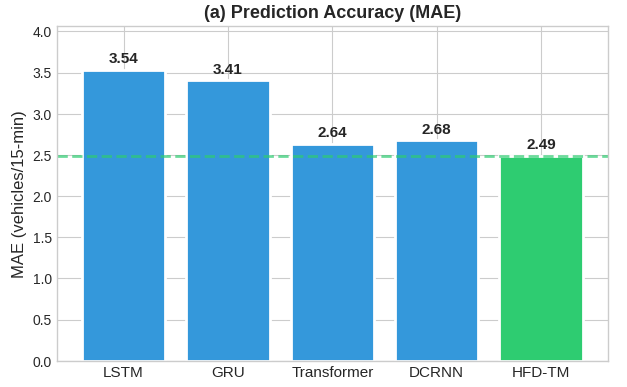}
    \caption{Comparison of mean absolute error (MAE) across baseline models and the proposed HFD-TM framework.}
    \label{fig:placeholder}
\end{figure}

\subsection{Theoretical Explanation for Error Reduction}

\label{sec:theory}
The observed error reductions are consistent with the statistical structure of intersection flows. Corridor through-movements dominate total volume (65.1\%) and exhibit lower volatility (mean coefficient of variation $\approx 0.80$) than turning movements (mean coefficient of variation $\approx 1.19$). Turning movements are moderately correlated with corridor totals (mean Pearson correlation $\approx 0.567$), and corridor totals explain a substantial portion of turning-movement variance (mean $R^2 \approx 0.355$). By the law of total variance,
\begin{equation}
\mathrm{Var}(Y_t) = \mathbb{E}\left[\mathrm{Var}(Y_t \mid Y_c)\right]
+ \mathrm{Var}\left(\mathbb{E}[Y_t \mid Y_c]\right),
\label{eq:total_variance}
\end{equation}
where $Y_t$ denotes a turning movement and $Y_c$ denotes corridor flow. Since $\mathrm{Var}\left(\mathbb{E}[Y_t \mid Y_c]\right)$ is non-trivial in this dataset, conditioning turn prediction on corridor flow reduces the residual uncertainty the model must learn. HFD-TM operationalizes this principle by first predicting lower-variance corridor flows and then estimating turning movements conditioned on these corridor predictions, yielding consistently lower prediction error than flat architectures.

\section{Conclusion}

This paper presents HFD-TM, a hierarchical deep learning framework for turning movement prediction at signalized corridor intersections. The framework decomposes the prediction task by first estimating stable corridor through-movements and then predicting turning movements conditioned on these estimates, while enforcing flow conservation through a physics-informed loss function.

Experimental results using real-world LiDAR data demonstrate that HFD-TM consistently outperforms GRU, LSTM, Transformer, and DCRNN baselines. Ablation analysis confirms that hierarchical decomposition is the primary contributor to performance improvement, consistent with observed traffic characteristics in which corridor flows exhibit lower volatility and explain a substantial portion of turning-movement variance. The framework also achieves significantly lower training time than graph-based baselines, supporting practical deployment.

Future work will extend the framework to corridors with diverse geometric and demand conditions, multi-step prediction horizons, and integration with signal timing optimization. Evaluation on additional corridors with varying geometric configurations is a natural next step to assess generalizability further. The proposed hierarchical formulation provides an efficient and principled approach for data-driven turning-movement prediction in intelligent transportation systems.




\bibliographystyle{IEEEtran}
\bibliography{references}

\end{document}